# SEMANTIC PROJECTION:
# RECOVERING HUMAN KNOWLEDGE OF MULTIPLE, DISTINCT OBJECT FEATURES FROM WORD EMBEDDINGS

**Gabriel Grand**[1*†], **Idan Blank**[2*], **Francisco Pereira**[3], and **Evelina Fedorenko**[4,5†]

**KEYWORDS:** word embeddings; distributional semantic models; lexical semantics; concepts; semantic features; word co-occurrence

## ABSTRACT

The words of a language reflect the structure of the human mind, allowing us to transmit thoughts between individuals. However, language can represent only a subset of our rich and detailed cognitive architecture. Here, we ask what kinds of common knowledge (semantic memory) are captured by word meanings (lexical semantics). We examine a prominent computational model that represents words as vectors in a multidimensional space, such that proximity between word-vectors approximates semantic relatedness. Because related words appear in similar contexts, such spaces – called "word embeddings" – can be learned from patterns of lexical co-occurrences in natural language. Despite their popularity, a fundamental concern about word embeddings is that they appear to be semantically "rigid": inter-word proximity captures only overall similarity, yet human judgments about object similarities are highly context-dependent and involve multiple, distinct semantic features. For example, dolphins and alligators appear similar in size, but differ in intelligence and aggressiveness. Could such context-dependent relationships be recovered from word embeddings? To address this issue, we introduce a powerful, domain-general solution: "semantic projection" of word-vectors onto lines that represent various object features, like size (the line extending from the word "small" to "big"), intelligence ("dumb" → "smart"), or danger ("safe" → "dangerous"). This method, which is intuitively analogous to placing objects "on a mental scale" between two extremes, recovers human judgments across a range of object categories and properties. We thus show that word embeddings inherit a wealth of common knowledge from word co-occurrence statistics and can be flexibly manipulated to express context-dependent meanings.

## 1. INTRODUCTION

When we say that we "know the meaning" of a word, what kind of knowledge do we mean we have? Words allow us to communicate the content of one human mind to another human mind: they are representations of mental structures. More precisely, we use words to express concepts (Marr, 1982). Concepts organize our knowledge about regularities in the world – they are generalizations about the kinds of things that exist and the properties that they have. Words, in turn, associate such abstract knowledge with surface forms of sounds/letters, and differences between word meanings correspond to distinctions that we can make between things in the world (Goldberg, 1995; Jackendoff, 2002; Murphy, 2004). Consequently, the psycholinguistic study of word meanings in the mental lexicon (lexical semantics; e.g., Steinberg and Jakobovits, 1971; Richards, 1982; Pinker and Levin, 1991) is tightly linked to the cognitive study of the architecture of world knowledge in the human mind (semantic memory; Quillian, 1966; Tulving, 1972).

---

[1] Department of Computer Science, Harvard University (ggrand@college.harvard.edu)
[2] McGovern Institute for Brain Research, Massachusetts Institute of Technology
[3] Medical Imaging Technologies, Siemens Healthineers
[4] Department of Psychiatry, Massachusetts General Hospital (evelina.fedorenko@mgh.harvard.edu)
[5] Department of Psychiatry, Harvard Medical School
\* *Equal contribution*
† *Corresponding author*

However, our world knowledge is broad, detailed and complex. Even an intuitively simple concept like DOG captures rich information about the animal it refers to, e.g., its appearance, biological properties, behavioral tendencies, cultural roles, etc. However, only a subset of this conceptual knowledge is communicated by the word "*dog*" (for a review, see Gleitman and Papafragou, 2013). Therefore, any theory of lexical semantics should specify the kinds of world knowledge that are captured in the lexicon. Yet constructing such a theory based on behavioral studies of language processing is challenging: it requires a paradigm for probing only that subset of conceptual distinctions that constitutes the (lexical) semantics of the word "*dog*" without also activating the rest of our world knowledge that is captured by the concept DOG.

Fortunately, a new insight into this problem is offered by computational methods that approximate a dissociation between lexical semantics and the rest of semantic memory: by granting a machine access only to word-forms, with no *a-priori* concepts, one can probe the semantic distinctions that are—in principle—available from the statistics of natural language alone. This approach relies on a hypothesis that dates back to the origins of modern linguistics, namely, that word meanings are determined by their patterns of usage, i.e., by the words they tend to appear with (De Saussure, 1916/2011; Wittgenstein, 1953/2010; Harris, 1954; Firth, 1957; Miller and Charles, 1991; for a review, see Sahlgren, 2008). Specifically, by tracking the distribution of word co-occurrences in multi-billion word corpora, unsupervised algorithms can learn a representation of word meanings as vectors in a multidimensional space where the proximity between these vectors increases with the co-occurrence probability of the corresponding words. The resulting space is called a "word embedding" or a "distributional semantic model" (for reviews, see Lenci, 2008; Turney and Pantel, 2010; Erk, 2012; Pereira et al., 2013; Baroni et al., 2014; Clark, 2015) (Bengio et al., 2003; Collobert et al., 2011; Huang et al., 2012).

Recent research has shown that inter-word distances in word embeddings correlate with human ratings of semantic similarity (Mikolov et al., 2013; Pennington et al., 2014). Furthermore, these distances are geometrically consistent across different word pairs that share a common semantic relation. For instance, the location of $\overrightarrow{man}$ relative to $\overrightarrow{woman}$ is similar to the location of $\overrightarrow{king}$ relative to $\overrightarrow{queen}$; This consistency allows for geometric operations to simulate conceptual relations, e.g., $\overrightarrow{king} - \overrightarrow{man} + \overrightarrow{woman} \approx \overrightarrow{queen}$ (Levy and Goldberg, 2014). These results demonstrate that, by simply tracking co-occurrence statistics, a machine with no *a-priori* concepts can obtain a lexicon that contains certain kinds of semantic memory.

Despite these impressive capabilities, word embeddings appear to have a fundamental limitation: the proximity between any two word-vectors captures only a single, semantically "rigid" measure of overall similarity. In contrast, humans evaluate the conceptual similarity between items in semantic memory flexibly, in a context-dependent manner. Consider, for example, our knowledge of dolphins and alligators: when we compare the two on a scale of size, from "small" to "big", they are relatively similar; in terms of their intelligence—on a scale from "dumb" to "smart"—they are somewhat different; and in terms of danger to us—on a scale from "safe" to "dangerous"—they differ significantly. Can such distinct, multiple relationships be inferred from word co-occurrence statistics? If so, how is such complex knowledge expressed in a word embedding?

Here, we suggest that such conceptual knowledge is implicitly embedded in the structure of word embeddings and we introduce a powerful, domain-general solution for extracting it: "semantic projection" of word-vectors onto "semantic subspaces" that represent different features. For instance, to recover the similarities in size among nouns in a certain category (e.g., $\overrightarrow{animals}$), we project their representations onto the line that extends from the word-vector $\overrightarrow{small}$ to the word-vector $\overrightarrow{big}$; to compare their levels of intelligence, we project them on the line connecting $\overrightarrow{dumb}$ and $\overrightarrow{smart}$; and to order them according to how dangerous they are, we project them onto the line connecting $\overrightarrow{safe}$ and $\overrightarrow{dangerous}$ (for an animation of this procedure, see: [Animated Semantic Projection Demo](#)). We demonstrate that the resulting feature-wise ("context-dependent") similarities robustly predict human judgments across a wide range of everyday object categories and semantic features. These results provide evidence that rich conceptual knowledge can be extracted bottom-up from the statistics of natural language, and it can be readily recovered from word embeddings with a simple and elegant method.

## 2. MATERIALS AND METHODS

Our experiment tested whether semantic projection (section 2.2.2) could recover context-dependent conceptual knowledge. To operationalize context-dependent knowledge, objects from a certain semantic category (e.g., animals) were rated based on different semantic features (e.g., size, intelligence, or danger). Namely, we chose several object categories along with semantic features that characterized them (section 2.1) and evaluated the ratings of category items along each feature in two ways: by collecting such ratings from humans, and via semantic projection in a word embedding that was created with Global Vectors for Word Representation (GloVe; Pennington et al., 2014). Our test then compared these two sets of feature-specific ratings.

2.1. SEMANTIC CATEGORIES AND FEATURES

We created a dataset of categories and features that met four criteria: (i) each category consisted of concrete objects/entities; (ii) each category was semantically associated with several features (to evaluate whether semantic projection generalized across features); (iii) each feature was semantically associated with several categories (to evaluate whether semantic projection generalized across categories); (iv) categories and features spanned diverse aspects of everyday semantic knowledge. All categories and their constituent items were selected from an extensive set of nouns, generated for a large-scale study on lexical memory (Mahowald et al., 2014).

2.1.1. CATEGORIES

Our dataset consisted of 9 semantic categories: animals, clothing, professions, weather phenomena, sports, mythological creatures, world cities, US states, and first names. These categories have been used in feature-elicitation studies, with varying degrees of prevalence (Paivio et al., 1968; Battig and Montague, 1969; Cree and McRae, 2003; McRae et al., 2005; Pereira et al., 2013; Binder et al., 2016): the first four have been frequently used; the next two—less frequently; and the last three have been rarely used but, unlike the other categories, consisted of proper nouns and, moreover, shared some features with the other categories. Items in all categories were used as cues in the word-association study of Nelson et al. (2004).[6]

Most categories consisted of 50 representative items (e.g., the clothing category contained words like "*hat*," "*tuxedo*," "*sandals*," and "*skirt*"; the full dataset is reproduced in the Appendix). The only exceptions were the categories of animals, which consisted of 34 items; weather, which consisted of 37 items; and professions, which consisted of 49 items. In order to ensure that items were representative of their respective categories, we chose within each category the 50 most frequent nouns according to the SubtlexUS Word Frequency Database (Brysbaert and New, 2009) (candidate items consisted of all the relevant nouns from Mahowald et al., 2014). We discarded multi-word expressions (except for US States, e.g., "*North-Dakota*") and words that did not appear in the vocabulary of the word embedding (section 2.2.1).

For two categories, additional selection criteria were used in order to increase the variability for the "gender" feature: for the first names category, we chose the 20 most common male names, 20 most common female names, and 10 most common unisex names from the past 100 years, based on public data from the US Social Security Administration.[7] For the professions category, efforts were made to balance the items by gender, because the majority of the most frequently occurring profession names were traditionally male.

2.1.2. FEATURES

Our dataset included 17 semantic features, each associated with a subset of the categories above: size, temperature, valence, loudness, speed, location, intelligence, wetness, weight, wealth, gender, danger, age, religiosity, political leaning, cost and arousal. These features have been produced in feature-elicitation studies with varying degrees of prevalence (Cree and McRae, 2003; McRae et al., 2005; Binder et al., 2016): the first four—almost invariably; the next three—frequently; the next six—less frequently; and the last four—very rarely. Words describing possible values of all features (e.g., "*big*" and "*small*" for the size feature), were used as cues in the word-association study of Nelson et al. (2004).

2.1.3. CHOOSING APPROPRIATE FEATURES FOR EACH CATEGORY

For each category in our dataset, only some features provided meaningful contexts for rating category members; for instance, sports could be rated by "danger", but not by "size", and names could be rated by "gender", but not "temperature". Therefore, out of the set of 9×17 = 153 category/feature pairings, we first selected a subset of 45 pairs that appeared intuitively appropriate. Then, to ensure that no appropriate pairs were mistakenly excluded, we collected human judgments about the relevance of each feature for each category. Specifically, we asked participants (*n*=50) on Amazon Mechanical Turk (MTurk, www.mturk.com; see Buhrmester et al., 2011) to rate how likely they were to describe each category (e.g., animals) in terms of each feature (e.g., size). Instead of using feature names, we used antonyms that represented opposite extreme values of that feature; for instance, a typical question read "How likely are you to describe animals as large/big/huge or small/little/tiny?". Ratings of the 153 category/feature pairs were made on a scale from 1 ("not likely at all") to 5 ("extremely likely").

---

[6] http://w3.usf.edu/FreeAssociation/
[7] http://www.ssa.gov/oact/babynames/decades/century.html; https://www.github.com/fivethirtyeight/data/tree/master/unisex-names

We averaged the ratings of each category/feature pair across participants. Out of the 153 pairs, we selected those for which the mean rating exceeded the 75$^{th}$ percentile (corresponding to a mean rating of 3.44 or above on the likelihood scale). This subset included 39 pairs. We removed two pairs where the norming question was misunderstood due to its ambiguous phrasing: animals-age (we are likely to describe an animal as being young or old, but comparing different species in terms of their age is less meaningful), and cities-political (we are likely to describe US cities as being democratic or republican, but many items included in this category for the main experiment were cities in other countries).

The remaining subset of 37 pairs partially overlapped with the manually pre-selected subset (27 pairs in common). We took the union of these two subsets to obtain a set of 56 pairs used in the main experiment. We note that the inclusion of pre-selected pairs that were not rated highly in the norming study could only weaken the performance of semantic projection: if semantic knowledge about these "unlikely" category/feature pairs is noisy or uncertain, we do not expect to recover it from a distributed semantic model (DSM) via semantic projection. In section 2.3, we describe how we measured this uncertainty in human knowledge (this resulted in the removal of 4 category/feature pairs for which such knowledge was extremely noisy).

## 2.2. SEMANTIC PROJECTION

### 2.2.1. THE GLOVE WORD EMBEDDING

We chose to conduct our experiment in the *GloVe* word embedding (Pennington et al., 2014), because it outperforms several other word embeddings in predicting word similarity judgments (Pereira et al., 2016). We used 300-dimensional GloVe vectors derived from the Common Crawl corpus, which contains approximately 42 billion uncased tokens and a total vocabulary size of 1.9 million.[8] In order to limit the vocabulary to words with robust co-occurrence estimates, we considered only the 500K most frequent words.

### 2.2.2. THE RATIONALE OF SEMANTIC PROJECTION

Semantic projection is a domain-general method for comparing word-vectors in the context of a certain semantic feature. A guiding example for applying this method in a simplified, three-dimensional GloVe space (for illustrative purposes) is depicted in **Figure 1** for the category "animals" and the feature "size". Intuitively, to compare animals in terms of this feature, we construct a scale—i.e., a straight line in GloVe space—on which animals could be ordered according to their size (the red line in **Figure 1**). This scale is constructed via a simple heuristic: we draw a line between antonyms—e.g., the word-vector $\overrightarrow{small}$ and the word-vector $\overrightarrow{large}$—that denote opposite values of the feature "size" (in **Figure 1**, these two word-vectors are denoted by red circles). This heuristic corresponds to taking a vector difference: "$\overrightarrow{size}$" $= \overrightarrow{large} - \overrightarrow{small}$ (we use quotation marks to distinguish between our scale, obtained by subtracting two word-vectors, and the vector of the lexical entry $\overrightarrow{size}$). Then, by projecting word-vectors of different animals onto this scale, we can approximate common knowledge about their relative sizes. For example, to estimate the relative size of a horse, we would compute the inner product $\overrightarrow{horse} \cdot$ "$\overrightarrow{size}$" (in **Figure 1**, this orthogonal projection is represented by the blue line extending from the blue dot of $\overrightarrow{horse}$ to the red scale of "$\overrightarrow{size}$").

The scale thus created is a one-dimensional subspace in which the feature "size" governs similarity patterns between word-vectors such that, for example, $\overrightarrow{horse}$ and $\overrightarrow{tiger}$ are located close to each other because they are relatively similar in size (in **Figure 1** these two word-vectors, denoted by blue circles, map onto nearby locations on the red line denoting the scale). Critically, these size-related similarity patterns might be different from the global similarities in the original space where, e.g., $\overrightarrow{horse}$ and $\overrightarrow{tiger}$ might be farther apart: despite their similarity in size (and other features), horses are perceived to be much less dangerous than tigers, they belong to a different taxonomic order, occupy different habitats, etc. (in **Figure 1**, the blue circle corresponding to $\overrightarrow{horse}$ is relatively far from that of $\overrightarrow{tiger}$, which happens to be closer to $\overrightarrow{rhino}$ and $\overrightarrow{alligator}$).

---

[8] https://nlp.stanford.edu/projects/glove/; https://www.commoncrawl.org/

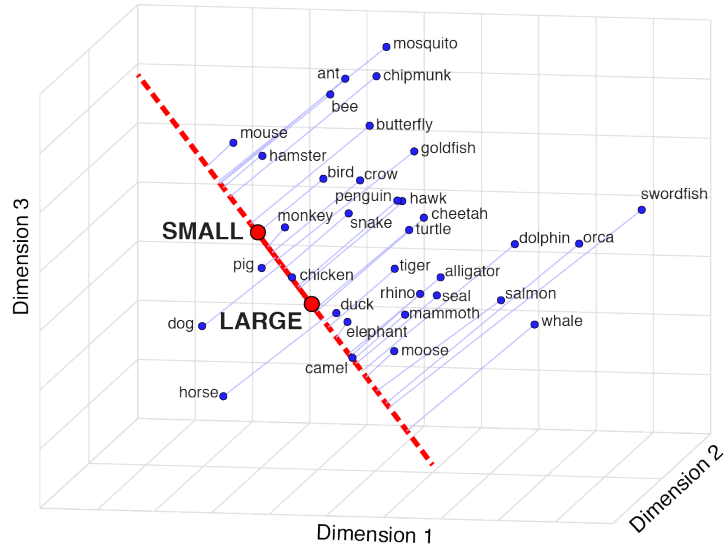

**Figure 1:** Schematic illustration of semantic projection. Word-vectors in the category "animals" (blue circles) are orthogonally projected (light blue lines) onto the feature subspace for "size" (red line), define as the vector difference between $\overrightarrow{large}$ and $\overrightarrow{small}$ (red circles). Dimensions are arbitrary and were chosen via Principal Component Analysis (PCA) to enhance visualization (the original GloVe word embedding has 300 dimensions). For an animated version of this figure, see: Animated Semantic Projection Demo.

Note that "size" is a semantic feature that applies to numerous categories of concrete objects: not only animals, but also mythological creatures, world cities, US states, etc. For each such category, its members could be projected onto the same size-scale described above. Hence, semantic projection on a "feature-subspace" is a domain-general method. In this article, we limit ourselves to semantic features that can be represented by one-dimensional subspaces ("scales"). However, other feature-subspaces for other semantic features could be of higher dimensionality (see Discussion).

### 2.2.3. DEFINING FEATURE SUBSPACES

A 1D feature subspace is approximated by the vector difference between antonyms that represent opposite ends of the feature continuum. Here, each end was represented not by a single word but, rather, by 3 words similar in meaning; the feature subspace was then defined as the average of the 3×3 = 9 possible vector differences (or "lines") between the two ends. For instance, the feature subspace "$\overrightarrow{size}$" was the average of the pairwise lines between $\{\overrightarrow{large}, \overrightarrow{big}, \overrightarrow{huge}\}$ and $\{\overrightarrow{small}, \overrightarrow{little}, \overrightarrow{tiny}\}$. This averaging procedure was chosen in order to obtain more robust approximations of feature subspaces that were not strongly impacted by the particular choice of antonyms. Indeed, lines for a given feature subspace were not strongly aligned, with the mean cosine similarity between one line and the average of the remaining eight being 0.533, corresponding to an angle of $0.99\pi$ (57 degrees). Nevertheless, lines were still more aligned within a feature subspace than across subspaces: the mean cosine similarity between one line and the average of lines from another subspace was 0.095, corresponding to an angle of $1.47\pi$ (84.5 degrees).

### 2.3. EVALUATING SEMANTIC KNOWLEDGE FROM HUMAN JUDGMENTS

Common knowledge about the 56 category/feature pairs in our dataset was evaluated via MTurk experiments, or "human information tasks" (HITs). These HITs were designed to be approximately 5 minutes long, and participants were paid $0.50 per HIT. In each HIT, $n$=25 participants rated the items in a single category according to a single feature. Each item had a separate sliding scale from 0 to 100, where 0 corresponded to a weak value of the feature (e.g., "small, little, or tiny" for size; "safe, harmless, or calm" for danger) and 100 corresponded to a strong value of the feature (e.g., "large, big, or huge" for "size"; "dangerous, deadly, or threatening" for danger). The words describing each end of the scale were the same words that we used to define the feature subspaces in the GloVe space.

In order to obtain data of high quality, we limited participation to MTurk users in the United Stated who had previously completed at least 1000 HITs with an acceptance rate of 95% or above. To account for participant idiosyncra-sies in the range of values used, we *z*-scored the ratings of each participant. To exclude participants who had rushed to complete their HIT or had otherwise provided random responses, we computed Pearson's moment correlation coefficient between the scaled responses of each participant and the average of the scaled responses across the rest of the sample. For each HIT, this procedure thus resulted in 25 inter-subject correlation values (*IS-rs*). The *IS-rs* for each HIT were Fisher-transformed in order to improve the normality of their distribution (Silver and Dunlap, 1987), and participants whose *IS-r* exceeded the mean *IS-r* in their sample by more than 2.5 standard deviations (i.e., participants whose ratings showed weak correlations with the rest of the group) were removed from further analysis. In the majority of HITs, no participants were excluded, and no more than 2 participants were excluded from any given HIT.

For a given category/feature pair, the average *IS-r* across participants—a measure of inter-rater reliability—provides a measure of the noise/uncertainty in common knowledge about that pair. Therefore, this measure estimates an upper bound for the fit between semantic projections and human data (section 2.4.2). When examining the average *IS-r* for each experiment, we identified four outlier experiments for which inter-rater reliability was very low (*IS-r*<0.07): cities by temperature, cities by wealth, clothing by arousal, and clothing by size. The *IS-r* values for these four experiments were clearly separate from the distribution of *IS-r* values across the remaining 52 experiments and, accordingly, we decided to remove them from further analysis. The same four experiments also appeared as outliers when we measured inter-rater reliability using a qualitatively different score (see section 2.4.2).

2.4. EVALUATING THE PREDICTIONS OF HUMAN JUDGMENTS GENERATED BY SEMANTIC PROJECTION

For each category/feature pair, we evaluated how well the ratings produced by semantic projection could recover the human ratings, by using two complementary measures.

*Linear correlation*. We used Pearson's moment correlation coefficient to estimate how much of the variance in human ratings across items could be predicted by variance in the ratings from the semantic projection. This measure is sensitive to even minor shifts in ratings: a slight change in a single item would (in most cases) affect the SD of the entire rating distribution and, consequently, change the contributions of all items to the correlation value. Therefore, it provides a rather strict test for our proposed method. Nonetheless, it is strongly biased by outliers, such that a strong correlation might reflect not the overall quality of semantic projection but, rather, a few extreme ratings made by both humans and our method.

*Pairwise order consistency*. This measure, which we denote $OC_p$, estimated the percentage of item pairs, out of all possible pairings, for which the difference in ratings had the same sign in both human judgments and the semantic projection. In other words, for every item pair (*i,j*) such that *i* was rated by humans as a having a stronger value than *j* on the feature continuum, we tested whether the semantic projection had predicted the same (vs. opposite) pattern. This measure is sensitive only to the direction of pairwise differences but not to their magnitude; for instance, in the animals/danger experiment, humans rate alligators to be more dangerous than dolphins, so we require that semantic projection makes the same judgment regardless of how far apart the two animals fall on the feature subspace. Here, a change in the rating of a single item would only affect those pairs that (a) include this item, and (b) reversed their difference sign as a result of this change. This measure is therefore robust to outliers.

2.4.1. SIGNIFICANCE TESTING

The significance of both evaluation measures for each experiment (i.e., category/feature pair) was measured via a permutation procedure. For each of 10,000 iterations, we randomly shuffled the labels of category items in the feature subspace (but not their labels in the human data) and recomputed our two evaluation measures to obtain their empirical null distributions. The significance of the true correlation was computed relative to the mean and SD of its null distribution, estimated with a Gaussian fit. To compute the significance of the real pairwise order consistency measure ($OC_p$), we counted the number of null values that exceeded it and divided this number by the total number of permutations. For each evaluation measure, *p*-values across the 52 experiments were corrected for multiple comparisons using False Discovery Rate correction (Benjamini and Yekutieli, 2001).

2.4.2. ESTIMATING AN UPPER BOUND FOR THE EVALUATION MEASURES

Because semantic projection approximates human knowledge, its success is limited by the amount of noise or uncertainty in that knowledge. Specifically, if human ratings for a certain category/feature pair exhibit low inter-rater reliability, then testing whether semantic projection captures human knowledge for this pair makes little sense, given that different people often disagree in their judgments. Therefore, for each experiment, we compared our first measure—linear correlation—to the reliability of ratings across participants, estimated via inter-subject correlation (*IS-r*;

see section 2.3). Specifically, we divided the squared correlation (i.e., percentage of variance in human ratings explained by semantic projection) by the squared *IS-r*, and took the square root of the result. We followed a similar procedure for computing pairwise order consistency across participants (inter-subject $OC_p$, or *IS-OC$_p$*), in order to obtain an upper bound for our second measure. Here, we divided the $OC_p$ from the semantic projection by the *IS-OC$_p$*. For both measures, values greater than 1 were set to 1. In the following sections, "adjustment for upper bound" refers to this normalization of our evaluation measures relative to inter-rater reliability of the behavioral data.

## 3. RESULTS

### 3.1. SEMANTIC PROJECTION PREDICTS HUMAN JUDGMENTS

For each of the 52 category/feature pairs, we tested how well the feature values of category items matched between human ratings and semantic projection. **Figure 2** shows scatterplots of human ratings against model predictions for each category/feature pair. **Figure 3** summarizes the distribution of the evaluation measures across all experiments, both before and after normalizing these measures relative to their upper bound (inter-rater reliability; section 2.4.2.).

Overall, semantic projection successfully recovered human semantic knowledge: moderate to strong correlations in ratings ($r>0.5$) were observed for nearly half of the pairs (25/52) and, across all experiments, the median correlation was 0.47. Moreover, after adjusting this statistic based on inter-rater reliability in ratings (median *IS-r* across experiments: 0.76), the "adjusted median correlation" was 0.61 (i.e., the variability in human ratings captured by semantic projection was 37% of the upper bound; $\sqrt{0.37} = 0.61$). Similarly, pairwise order consistency between semantic projection and human ratings was higher than 50% for nearly all pairs (47/52) and, across all experiments, the median $OC_p$ was 65%. After adjusting $OC_p$ based on inter-rater reliability (median *IS-OC$_p$* across experiments: 73%), the median increased to 90% (i.e., the pairwise order consistency in human ratings captured by semantic projection was 90% of the upper bound). In about half of the experiments (31/52), both the correlation and $OC_p$ measures were significant.

However, the fit between semantic projection and human ratings varied a lot across experiments, with the best fit observed for ratings of names by gender ($r=0.94$, $OC_p=87\%$) and the worst fits observed for ratings of cities by cost ($r=-0.15$, $OC_p=47\%$) and professions by location ($r=-0.12$, $OC_p=45\%$) (Inter-quartile range across experiments: $r=0.44$, $OC_p=16\%$). We note that across the 52 experiments, our evaluation measures did not correlate strongly with their corresponding reliability measures (i.e., upper bounds). Specifically, the correlation between $r$ and *IS-r* across these experiments was 0.22, and the correlation between $OC_p$ and *IS-OC$_p$* was 0.09 (both are not significant, as indicated by 5000 random permutations of experiment labels). We further consider this variability across experiments in the Discussion section.

### 3.2. REMOVING ONE END OF A FEATURE SCALE COMPROMISES SEMANTIC PROJECTION

In defining a feature subspace (e.g., $\overrightarrow{"size"}$) for semantic projection, we relied on antonymous adjectives (e.g., $\overrightarrow{small}$ and $\overrightarrow{big}$) that can be intuitively thought of as defining opposite ends of a scale. The arithmetic difference between the antonymous word vectors situates this scale within the GloVe space, i.e., it defines a "diagnostic" direction in the space along which concrete objects show high variation with regard to the feature in question. However, this subtraction might not be necessary if the diagnostic information were already sufficiently represented by each adjective vector on its own. In other words, the word-vector $\overrightarrow{big}$ might already define a size-related diagnostic direction in GloVe space, such that subtracting $\overrightarrow{small}$ from it is redundant (this would be possible if, e.g., these antonyms lay on opposite sides of the GloVe origin). To test whether one end of a feature scale was sufficient for predicting human common knowledge, we repeated all experiments but this time compared behavioral ratings to (i) semantic projection on the end that represented "more" of a feature (e.g., the average of $\{\overrightarrow{large}, \overrightarrow{big}, \overrightarrow{huge}\}$ for "size"); and, separately, (ii) semantic projection on the end that represented "less" of a feature (e.g., $\{\overrightarrow{small}, \overrightarrow{little}, \overrightarrow{tiny}\}$ for "size").

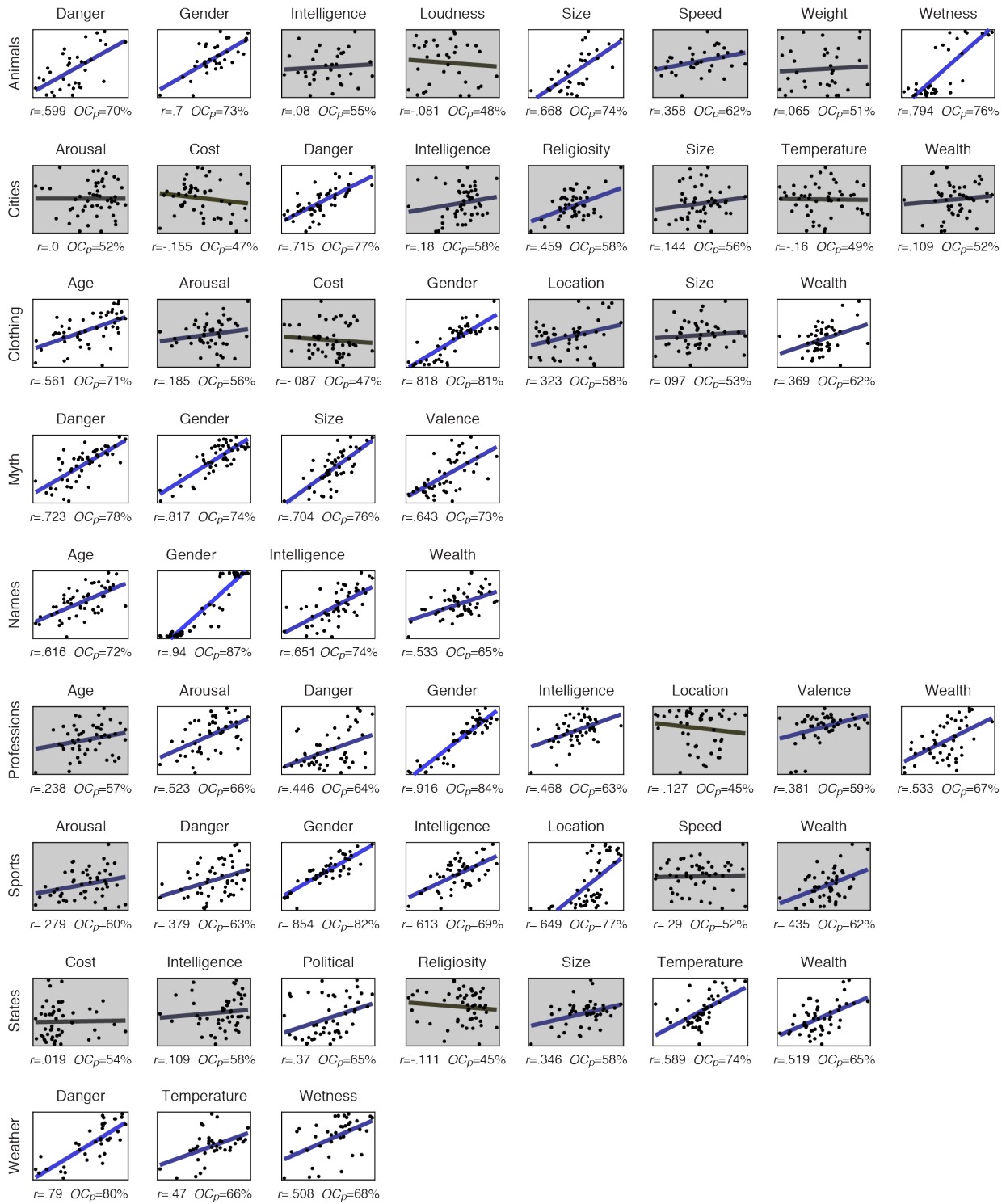

**Figure 2:** Semantic projection predicts human judgments across various categories and features. For each of 52 category/feature pairs, scatterplots show the relationship between *z*-scores of average item ratings across participants (y-axis) and ratings predicted by semantic projection (*x*-axis). Correlation and pairwise order consistency value are presented below each scatterplot. Experiments for which both of these measures were significant are shown over a white background. Straight lines are linear regression fits to the data and, across figures, vary according to correlation strength from black (weak) to blue (strong).

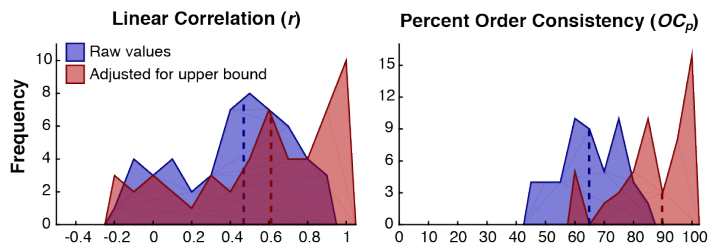

**Figure 3:** Distribution of evaluation scores for semantic projection. Histograms across 52 category/feature pairs are presented for linear correlations (left) and pairwise order consistency (right) between semantic projection ratings and human judgments, before (blue) and after (red) adjustment for inter-participant reliability in ratings. Dashed lines denote medians.

Across the 52 experiments, projections on the former ("more") end had a median correlation of $r=0.18$ with human ratings, and projections on the latter ("less") end had a median correlation of $r=0$; both projection schemes performed worse than our original semantic projection ($p<10^{-6}$, $d=0.75$ and $p<10^{-8}$, $d=0.99$, respectively) which, as reported above, had a median correlation of $r=0.47$ with human ratings. Similar patterns emerged for the other evaluation score, namely, pairwise order consistency: across experiments, the two alternative projection schemes had median values of $OC_p=58\%$ and $OC_p=50\%$, both worse than our original semantic projection ($p<10^{-4}$, $d=0.58$ and $p<10^{-7}$, $d=0.86$, respectively), which had a median value of $OC_p=66\%$. Performance was even worse when, instead of using projection, we computed the distance between each item in a category and either end of the feature scale, using either cosine or Euclidean distance. Therefore, the difference between the vectors of antonymous adjectives—rather than either vector in isolation—provides feature-specific "diagnostic" dimension in GloVe space (see also Kozima and Ito, 1995).

### 3.3. SEMANTIC PROJECTION IS SUCCESSFUL EVEN WITHOUT OUTLIER ITEMS

One potential concern that arose during our analysis was that the correlation between human ratings and semantic projection might have resulted from few outliers that were rated as having extreme feature values by both humans and our method. For instance, when rating animals by size, semantic projection might be able to predict that whales are very big while mice are very small based on word co-occurrence statistics: the words "*big*", "*large*" or "*huge*" (which define one end of the size subspace) might co-occur with some frequency around "*whale*", whereas the words "*small*", "*little*" or "*tiny*" (which define the other end of the size subspace) might co-occur with some frequency around "*mouse*". Such extreme items could cause a strong fit between semantic projection and human ratings even if, for most animals with less extreme values, semantic projection cannot recover human knowledge. However, we do not believe this to be the case, because semantic projection showed high pairwise order consistency with human ratings, and this measure is not strongly biased by outliers.

Still, in order to address this concern more directly, we repeated our analyses for each of the 31 significant experiments after removing the item that received the most extreme average rating, as measured by the absolute value of its z-score across items; then, we removed the next most extreme item, repeated the analyses, and continued this process until 10 items (i.e., 20% of the 50-item categories) had been removed. The results of this analysis are presented in **Figure 4**. As extreme items were gradually removed from the dataset, both evaluation measures somewhat decreased: across experiments, the median initial correlation of 0.62 decreased to 0.50 after removing the 10 most extreme items. Similarly, the median initial $OC_p$ of 73%, decreased to 66%. However, removing extreme items also resulted in less reliable human ratings (reflecting relatively higher noise or uncertainty for less extreme items): across experiments, the median *IS-r* decreased from 0.80 to 0.66, and the median *IS-OC_p* decreased from 75% to 69%.

To test whether our evaluation measures were more severely affected by the removal of extreme items, compared to the inter-participant reliability measures, we carried out a mixed-effects linear regression to predict the values of both measures based on: (i) the number of removed items; (ii) the type of measure (raw vs. upper-bound approximation); and (iii) the interaction between (i) and (ii). Random intercepts by category and by feature were included. For both correlation and $OC_p$, the interaction term did not significantly improve the model (modeling $r$ and *IS-r*: $\chi^2_{(1)}=0.10$, $p=0.75$; modeling $OC_p$ and *IS-OC_p*: $\chi^2_{(1)}=0.22$, $p=0.64$), indicating that the decrease in the evaluation measures was indistinguishable from the decrease in the corresponding inter-participant reliability. Therefore, extreme feature values bias the success of semantic projection only to the same extent that they increase certainty, or reduce noise, in semantic knowledge itself (as reflected by increased reliability of human judgments).

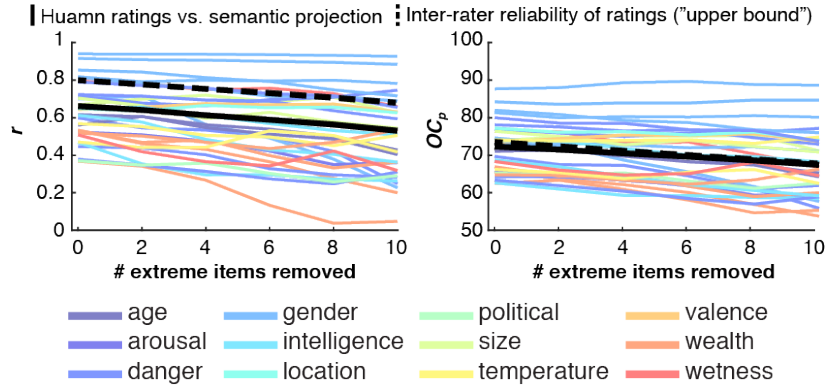

**Figure 4:** Semantic projection is not biased by outliers. r (left) and OCp (right) values are plotted as a function of the number of items with the most extreme human ratings that were removed for the data. Data from each of 31 category/feature pairs (those with significant results in our main experiment) are depicted in color, varying by feature. Black continuous lines show the average across the 31 pairs. Black dotted lines show the average inter-rater reliability across the 31 pairs.

## 4. DISCUSSION

Our findings demonstrate that semantic projection of concrete nouns can approximate human ratings of the corresponding entities along multiple, distinct feature continuums. The method we introduce is simple, yet robust, successfully predicting human judgments across a range of everyday object categories and semantic features. These results suggest that semantic knowledge about context-dependent similarities is implicitly represented in the structure of word embedding spaces. Thus, the conceptual knowledge that can be constructed bottom-up from word co-occurrence statistics is significantly more flexible and rich than has been previously assumed.

The nature of semantic knowledge representations has been long debated, and continues to be central to the study of the human mind in cognitive science, neuroscience and philosophy (Jackendoff, 1992; Saffran and Schwartz, 1994; Fodor, 1998; Laurence and Margolis, 1999; Barsalou, 2008; Lambon-Ralph and Patterson, 2008; Mahon and Caramazza, 2008; Binder et al., 2009; Markman, 2013; Yee et al., 2014; Margolis and Laurence, 2015; Mahon and Hickok, 2016). In particular, knowledge about the relationships between object categories and the features characterizing them has been traditionally discussed in the context of symbolic representations such as feature lists (Rosch and Mervis, 1975; Smith and Medin, 1981), structured schemata (Rumelhart and Ortony, 1977), or highly elaborate intuitive theories (Gopnik et al., 1997; Gopnik, 2003) (for an early example of modeling context-dependent similarities, see Medin and Schaffer, 1978). Prior studies that have attempted to extract such knowledge from natural corpora have had to augment the tracking of word co-occurrences with more elaborate information such as dependency parses or supervised identification of particular linguistic patterns (Poesio and Almuhareb, 2005; Barbu, 2008; Baroni and Lenci, 2009; Baroni et al., 2010; Kelly et al., 2014; Rubinstein et al., 2015). Our results therefore challenge these approaches by suggesting that the distributional, sub-symbolic representational format of word embeddings can support the flexible re-structuring of object categories to reflect specific features. Moreover, given that these spaces are hypothesized to approximate lexical semantics, the current study suggests that complex feature-specific knowledge is part of a word's meaning.

In addition to demonstrating that word meanings may integrate knowledge that had been independently acquired through non-linguistic (e.g., perceptual) experience, our findings provide a proof-of-principle that such knowledge can be independently acquired from statistical regularities in natural language itself. In other words, the current study is consistent with the intriguing hypothesis that, like word embedding spaces, humans can use language as a gateway to acquiring conceptual knowledge (Rumelhart, 1979; Landauer and Dumais, 1997; Elman, 2004, 2009; Lupyan and Bergen, 2016). Indeed, humans are sensitive to patterns of word co-occurrence, and use them during language processing (McDonald and Ramscar, 2001; Ellis and Simpson-Vlach, 2009; Frank and Thompson, 2012; Monsalve et al., 2012; Smith and Levy, 2013; Willems et al., 2015) (for tracking of relationships between words and linguistic contexts more generally, see, e.g., Clifton et al., 1984; MacDonald et al., 1994; Trueswell et al., 1994; Garnsey et al., 1997; Hale, 2001; Traxler et al., 2002; Gennari and MacDonald, 2008; Levy, 2008). In addition, evi-dence from congenitally blind individuals suggests that such patterns are indeed sufficient for acquiring some forms of perceptual knowledge, e.g., similarities between colors or actions involving motion, and subtle distinctions between sight-verbs such as "*look*", "*see*" and "*glance*" (Marmor, 1978; Shepard and Cooper, 1992; Noppeney et al., 2003; Landau et al., 2009; Bedny et al., 2011). Thus, in the absence of direct, perceptual experience, language itself can serve as a source of semantic knowledge.

What kinds of category/feature relations could be acquired in this manner? Past research has suggested word embedding spaces (prior to any semantic projection) capture gross knowledge about the sensory modalities associated with different objects (Louwerse and Connell, 2011), but they fare relatively poorly in approximating detailed perceptual properties in comparison to abstract (e.g., encyclopedic or functional) knowledge (Baroni and Lenci, 2008; Andrews et al., 2009; Baroni et al., 2010; Riordan and Jones, 2011; Hill et al., 2016). The current results suggest a more nuanced view (see **Figure 2**): first, knowledge about some perceptual features (e.g., size) was successfully predicted for some categories (animals, mythological creatures). Second, whereas some abstract features (e.g., gender, danger) could be recovered via semantic projection across all the categories with which they were paired, other abstract features (e.g., intelligence) were only recovered for some categories but not others. The factors underlying the performance patterns of semantic projection across different feature/category pairs (and across different entities within a category) may, therefore, be a fruitful area of future investigation. For instance, a recent study has reported a correspondence between implicit social biases and the knowledge available in word embeddings (Caliskan et al., 2017).

Notwithstanding this variability in performance, we emphasize that semantic projection exhibits promising generalizability across animate and inanimate categories (e.g., animals vs. clothing), natural and man-made categories (e.g., weather vs. mythological creatures), and common and proper nouns (e.g., professions vs. cities). We are excited to further test how well this method can recover knowledge about entities that are abstract rather than concrete, as well as knowledge about concepts that correspond to parts of speech other than nouns (e.g., projecting verbs on adverb-subspaces). Similarly, our method generalizes across different "kinds" of features: from those that are judged to be relatively binary (e.g., the wetness of animals) to those that vary more continuously between two extremes (e.g., the gender associated with articles of clothing). It may further extend to other kinds of features such as those with multiple, discrete values (e.g., "number of legs") and, more generally, to complex types of context-dependent knowledge represented in semantic subspaces with more than one dimension.

These prospects for generalizability raise deeper questions about semantic knowledge representations in word embeddings: which set of word-vectors are considered to constitute psychologically plausible categories (Murphy, 2004)? Which semantic subspaces represent features (or, more generally, contexts; Clark, 1973; Binder et al., 2016)? Which categories can be meaningfully described with which features (e.g., Barsalou and Sewell, 1985; Tanaka and Taylor, 1991)? And what geometric operations besides than linear projection could predict different kinds of human knowledge? Addressing these questions could help characterize the structure of word embedding spaces and, critically, inform more general theories of categories and features that are fundamental to the study of concepts. Specifically, if word embeddings are found to implicitly represent information that approximates human patterns of category formation, feature elicitation, and context-dependent semantic judgments, then their structure could perhaps provide a principled way for deriving an ontology of concepts.

In conclusion, semantic projection in word embeddings is a powerful method for estimating human knowledge about the structure of categories under distinct contexts. Within the distributional semantics literature, our method continues the tradition of applying simple linear algebraic operations to perform useful semantic comparisons in word embeddings (e.g., vector subtraction, cosine similarities, matrix multiplication; Baroni and Zamparelli, 2010; Mikolov et al., 2013). Moreover, compared to prior attempts at extracting semantic knowledge from patterns of natural language use, this method requires significantly less human supervision and/or corpora annotation. Most importantly, it obviates the need to define *a-priori* a constrained ontology of semantic features that would span the vector space: given an existing word embedding, semantic projection can flexibly recover a variety of semantic features. Therefore, we believe that we have only scratched the surface of the total volume of knowledge implicit in word embeddings. We hope that semantic projection will provide a useful, generalizable framework for deeper exploration of such models.

APPENDIX: CATEGORY ITEMS AND FEATURE DEFINITIONS

1. CATEGORY ITEMS

*1.1. Animals (n=34)*
Alligator, Ant, Bee, Bird, Butterfly, Camel, Cheetah, Chicken, Chipmunk, Crow, Dog, Dolphin, Duck, Elephant, Goldfish, Hamster, Hawk, Horse, Mammoth, Monkey, Moose, Mosquito, Mouse, Orca, Penguin, Pig, Rhino, Salmon, Seal, Snake, Swordfish, Tiger, Turtle, Whale

*1.2. Cities (n=50)*
Amsterdam, Atlanta, Baghdad, Bangkok, Barcelona, Beijing, Berlin, Boston, Cairo, Chicago, Dallas, Delhi, Detroit, Dubai, Hong-Kong, Honolulu, Houston, Istanbul, Jakarta, Jerusalem, Johannesburg, Karachi, Kiev, Lahore, London, Los-Angeles, Madrid, Mexico, Miami, Milan, Montreal, Moscow, Mumbai, Nairobi, New-York, Paris, Philadelphia, Prague, Rome, San-Francisco, Seoul, Shanghai, Singapore, Taipei, Tehran, Tokyo, Toronto, Venice, Vienna, Warsaw

*1.3. Clothing (n=50)*
Bathrobe, Belt, Bikini, Blouse, Boots, Bra, Bracelet, Coat, Collar, Cuff, Dress, Earrings, Glasses, Gloves, Gown, Hat, Jacket, Jeans, Knickers, Loafers, Necklace, Nightgown, Overcoat, Pajamas, Panties, Pants, Pantyhose, Raincoat, Robe, Sandals, Scarf, Shawl, Shirt, Shorts, Skirt, Sleeve, Slippers, Sneakers, Socks, Stockings, Sweater, Sweatshirt, Swimsuit, Thong, Tiara, Tights, Trousers, Tuxedo, Vest, Watch

*1.4. Mythological creatures (n=50)*
Alien, Angel, Banshee, Bigfoot, Centaur, Chimera, Cyclops, Demon, Devil, Dragon, Dwarf, Elf, Fairy, Faun, Genie, Ghost, Ghoul, Gnome, Goblin, God, Golem, Gorgon, Gremlin, Griffin, Harpy, Hydra, Imp, Leprechaun, Leviathan, Manticore, Mermaid, Minotaur, Nymph, Ogre, Pegasus, Phoenix, Poltergeist, Sphinx, Succubus, Thunderbird, Titan, Troll, Unicorn, Vampire, Warlock, Werewolf, Witch, Wizard, Wraith, Zombie

*1.5. First names (n=50)*
Anthony, Ashley, Avery, Barbara, Betty, Carol, Casey, Charles, Christopher, Daniel, David, Donald, Donna, Dorothy, Elizabeth, George, Jackie, Jaime, James, Jennifer, Jessica, Jessie, Jody, John, Joseph, Karen, Kendall, Kenneth, Kerry, Kimberly, Linda, Lisa, Margaret, Mark, Mary, Matthew, Michael, Nancy, Patricia, Paul, Peyton, Richard, Riley, Robert, Sandra, Sarah, Steven, Susan, Thomas, William

*1.6. Professions (n=49)*
Actor, Actress, Artist, Attorney, Babysitter, Boss, Businessman, Businesswoman, Carpenter, Chef, Coach, Cop, Dancer, Detective, Doctor, Driver, Firefighter, Fool, Gardener, Guard, Housekeeper, Janitor, Judge, King, Lawyer, Librarian, Lieutenant, Maid, Mailman, Manager, Mechanic, Nurse, Pilot, President, Prince, Princess, Prisoner, Professor, Psychiatrist, Queen, Secretary, Sheriff, Soldier, Student, Teacher, Thief, Waiter, Waitress, Writer

*1.7. Sports (n=50)*
Aerobics, Archery, Badminton, Baseball, Basketball, Boxing, Canoeing, Cheerleading, Curling, Cycling, Dancing, Diving, Fencing, Fishing, Football, Golf, Gymnastics, Handball, Hiking, Hockey, Hunting, Jogging, Karate, Kayaking, Lacrosse, Marathon, Ping-Pong, Racing, Racquetball, Rollerblading, Rowing, Rugby, Running, Sailing, Skateboarding, Skiing, Skydiving, Sledding, Snorkeling, Snowboarding, Soccer, Softball, Sumo, Surfing, Swimming, Tennis, Volleyball, Weightlifting, Windsurfing, Wrestling

*1.8. US states (n=50)*
Alabama, Alaska, Arizona, Arkansas, California, Colorado, Connecticut, Delaware, Florida, Georgia, Hawaii, Idaho, Illinois, Indiana, Iowa, Kansas, Kentucky, Louisiana, Maine, Maryland, Massachusetts, Michigan, Minnesota, Mississippi, Missouri, Montana, Nebraska, Nevada, New-Hampshire, New-Jersey, New-Mexico, New-York, North-Dakota, North-Carolina, Ohio, Oklahoma, Oregon, Pennsylvania, Rhode-Island, South-Carolina, South-Dakota, Tennessee, Texas, Utah, Vermont, Virginia, Washington, West-Virginia, Wisconsin, Wyoming

*1.9. Weather phenomena (n=37)*
Blizzard, Breeze, Cloud, Cyclone, Dew, Drizzle, Drought, Earthquake, Flood, Flurry, Fog, Frost, Hail, Heatwave, Humidity, Hurricane, Lightning, Mist, Monsoon, Rain, Rainbow, Sandstorm, Sleet, Smog, Snow, Squalls, Storm, Sunshine, Thunder, Thunderstorm, Tides, Tornado, Tsunami, Twister, Typhoon, Whirlwind, Wind

2. FEATURE DEFINITIONS

| | | | |
|---|---|---|---|
| Age: | {old, ancient, elderly} | ↔ | {young, youth, child} |
| Arousal: | {interesting, exciting, fun} | ↔ | {boring, unexciting, dull} |
| Cost: | {expensive, costly, fancy} | ↔ | {inexpensive, cheap, budget} |
| Danger: | {dangerous, deadly, threatening} | ↔ | {safe, harmless, calm} |
| Gender: | {male, masculine, man} | ↔ | {female, feminine, woman} |
| Intelligence: | {intelligent, smart, wise} | ↔ | {stupid, dumb, idiotic} |
| Location: | {indoor, indoors, inside} | ↔ | {outdoor, outdoors, outside} |
| Loudness: | {loud, deafening, noisy} | ↔ | {soft, silent, quiet} |
| Political: | {democrat, liberal, progressive} | ↔ | {republican, conservative, redneck} |
| Religiosity: | {religious, spiritual, orthodox} | ↔ | {atheist, secular, agnostic} |
| Size: | {large, big, huge} | ↔ | {small, little, tiny} |
| Speed: | {fast, speedy, quick} | ↔ | {slow, sluggish, gradual} |
| Temperature: | {hot, warm, tropical} | ↔ | {cold, cool, frigid} |
| Valence: | {good, great, happy} | ↔ | {bad, awful, sad} |
| Wealth: | {rich, wealthy, privileged} | ↔ | {poor, poverty, underprivileged} |
| Weight: | {heavy, fat, thick} | ↔ | {light, skinny, thin} |
| Wetness: | {wet, water, ocean} | ↔ | {dry, country, land} |